\documentclass[letterpaper, 10 pt, conference]{ieeeconf}

\IEEEoverridecommandlockouts                              %

\usepackage{cite}
\usepackage{amsmath,amssymb,amsfonts}
\usepackage{algorithmic}
\usepackage{graphicx}
\usepackage{subfigure}
\usepackage{textcomp}
\usepackage{xcolor}
\usepackage{multirow} 
\usepackage{multicol} 
\usepackage[bookmarks=false]{hyperref}

\newcommand{\fig}[1]{Fig.~\ref{fig:#1}}

\def\BibTeX{{\rm B\kern-.05em{\sc i\kern-.025em b}\kern-.08em
    T\kern-.1667em\lower.7ex\hbox{E}\kern-.125emX}}

\title{\LARGE \bf
Efficient Fully Convolution Neural Network for Generating Pixel Wise Robotic Grasps With High Resolution Images 
}

\author{Shengfan Wang$^{\dagger}$, Xin Jiang$^{*}$, Jie Zhao, Xiaoman Wang, Weiguo Zhou and Yunhui Liu, Fellow, IEEE%
\thanks{This work was supported by the following projects: Shenzhen Peacock Plan Team grant (KQTD20140630150243062), Shenzhen and Hong Kong Joint Innovation Project (SGLH20161209145252406), Shenzhen Fundamental Research grant (JCYJ20170811155308088).
}%
\thanks{Shengfan Wang, Xin Jiang, Jie Zhao, Xiaoman Wang and Weiguo Zhou are with the School of Mechanical Engineering and Automation, Harbin Institute of Technology, Shenzhen 518055, China. The author e-mail: 18S053234@stu.hit.edu.cn. The corresponding author email: x.jiang@ieee.org.}%
\thanks{Yunhui Liu is with the Department of Mechanical and Automation Engineering, The Chinese University of Hong Kong, Shatin, Hong Kong, China.}%
}

\begin{document}
\maketitle

\begin{abstract}
This paper presents an efficient neural network model to generate robotic grasps with high resolution images. The proposed model uses fully convolution neural network to generate robotic grasps for each pixel using 400 $\times$ 400 high resolution RGB-D images. It first down-sample the images to get features and  then up-sample those features to the original size of the input as well as combines local and global features from different feature maps. Compared to other regression or classification methods for detecting robotic grasps, our method looks more like the segmentation methods which solves the problem through pixel-wise ways. We use Cornell Grasp Dataset to train and evaluate the model and get high accuracy about 94.42\% for image-wise and 91.02\% for object-wise and fast prediction time about 8ms. We also demonstrate that without training on the multiple objects dataset, our model can directly output robotic grasps candidates for different objects because of the pixel wise implementation.
\end{abstract}

\section{Introduction}
Researchers have spent large amount of time trying to solve the grasp problem in Robotics. While human beings can easily grasp any object around them in multiple ways, robots still cannot for various reasons related to vision and planning. A robot mush know where a object is first and then determine the pose of its gripper to grasp the object. We treat the above problem as a grasp detection problem and try to solve it by using vision, especially through RGB-D cameras.

Previous works treat the grasp detection either as a classification\cite{5980145} problem or as a regression \cite{7139361} problem. For classification methods, they usually detect the grasp first by using methods like sliding window to search the potential grasp space\cite{Lenz2013Deep} and then using neural networks to rank them separately\cite{mahler2017dex}, which is time-consuming for the complex procedure. For regression methods, they tend to use neural networks to output the coordinates of the grasps directly \cite{Kumra2016Robotic}. However, for the property of regression, these methods will output the average of the ground truth grasps, which may lead to unreasonable grasps. 

\begin{figure}[tb] 
\centering 
\subfigure[one grasp candidate]{ \includegraphics[width=3.6cm]{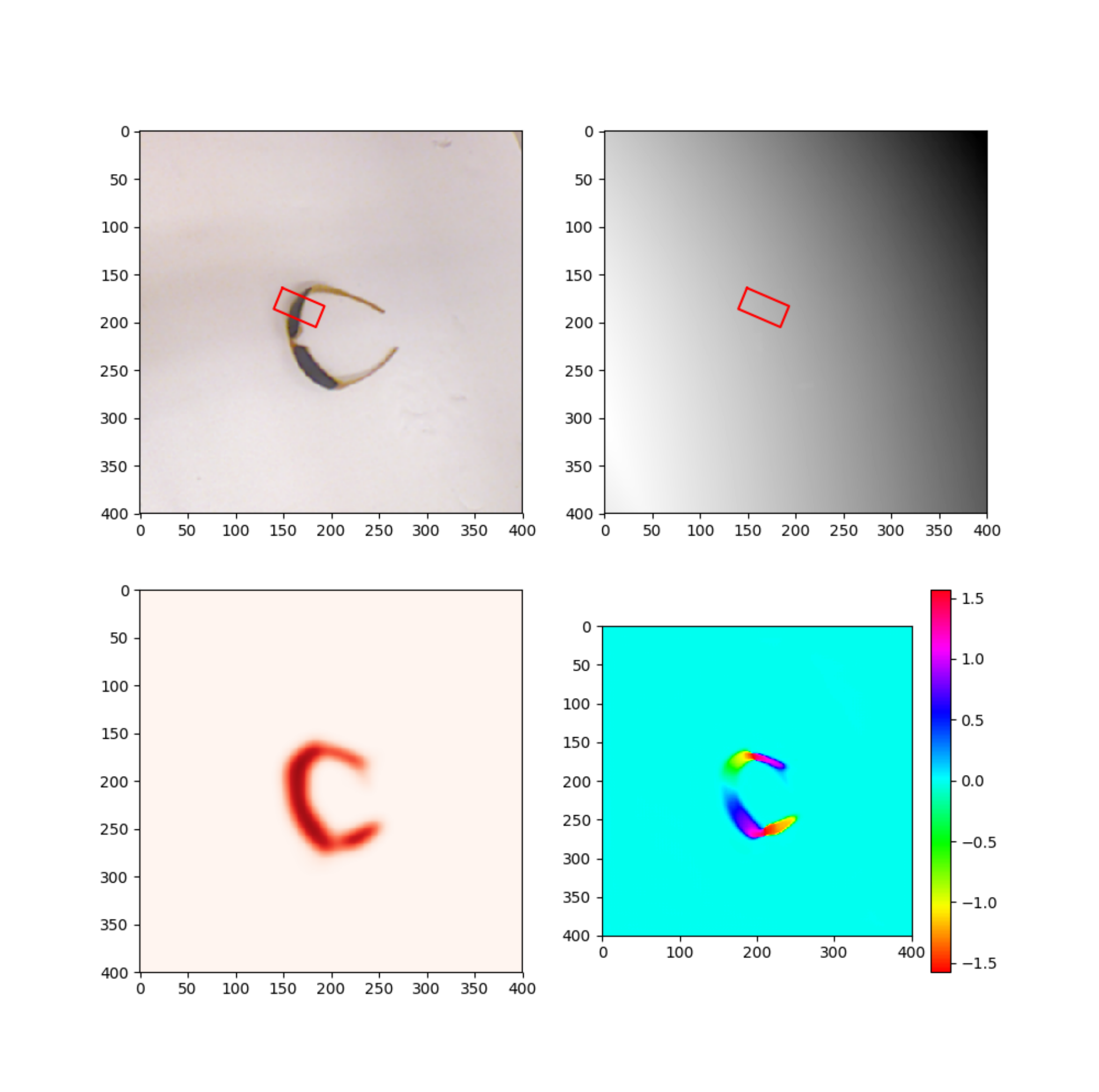} } 
\quad 
\subfigure[two grasp candidates]{ \includegraphics[width=3.6cm]{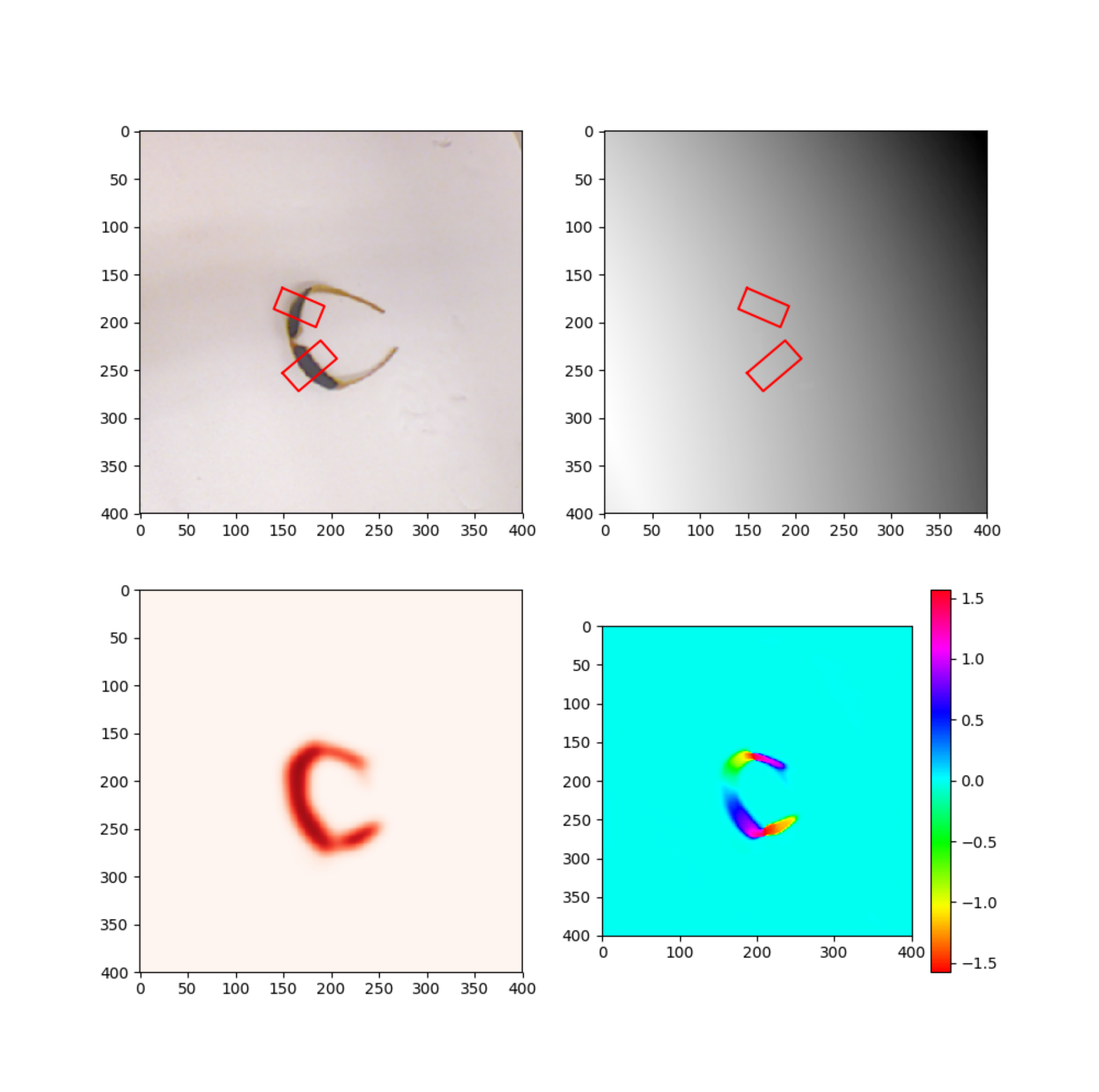} } \quad 
\subfigure[three grasp candidates]{ \includegraphics[width=3.6cm]{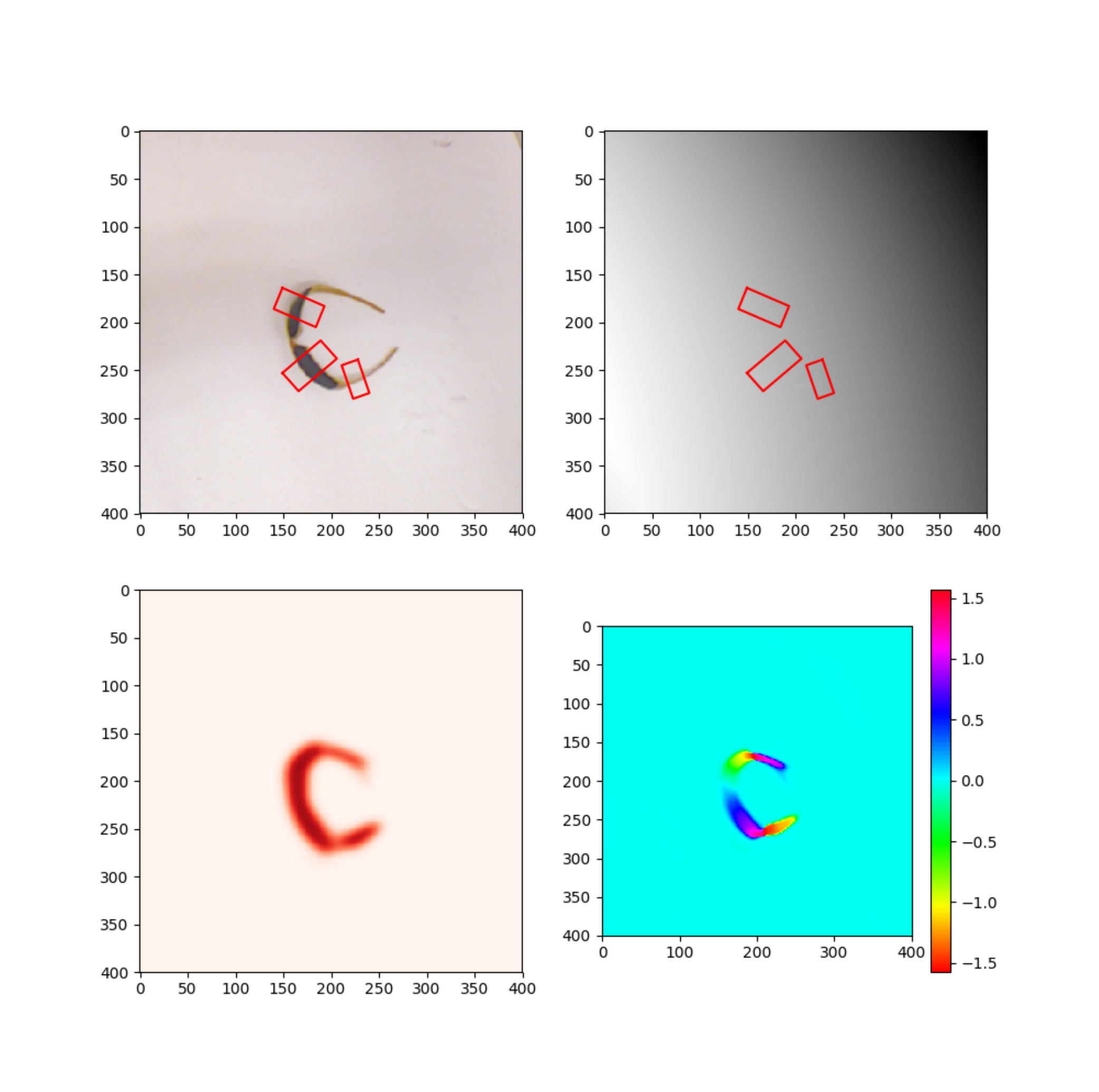} } \quad 
\subfigure[four grasp candidates]{ \includegraphics[width=3.6cm]{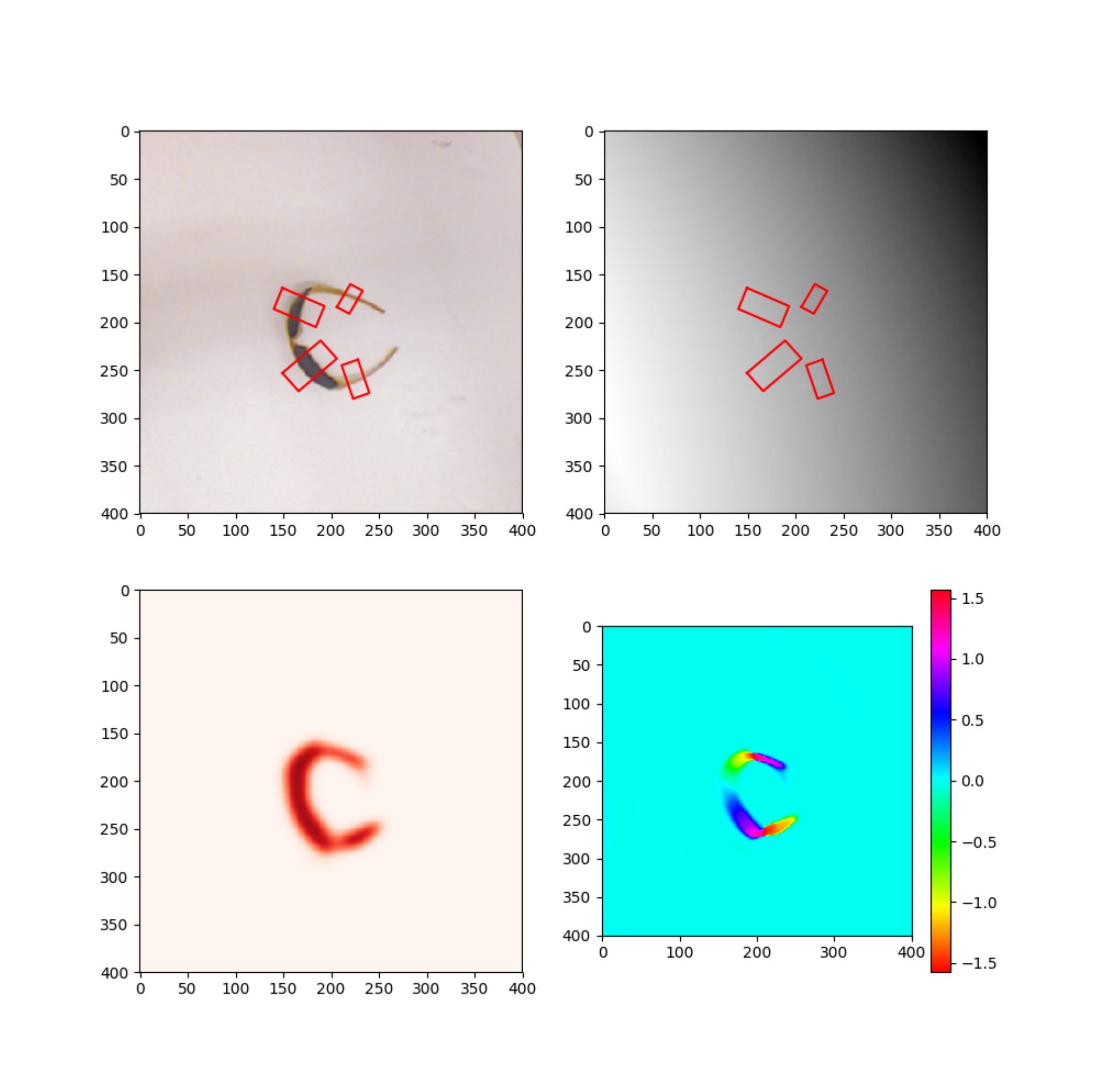} } 
\caption{ Robotic grasps predicted by our proposed model. Top left is the raw color image. Top right is the raw depth image. Bottom left is the grasp position prediction. Bottom right is the grasp angle prediction.} 
\label{fig:demo}
\end{figure}

Our proposed method tries to solve the problem utilizing some ideas from segmentation tasks \cite{Long2015Fully} and was inspired by the GG-CNN \cite{Morrison2018Closing}. Instead of evaluating whole grasp candidates to find the best grasps in RGB-D images like classification methods or approximating the 2-D grasp localization and orientation for each object like regression methods, our method predicts pixel-level robotic grasp candidates for objects through one forward propagation. Moreover, recent works \cite{Chu2018Deep,8403246,zhang2018roi} try to output multiple grasp candidates by utilizing skills from object detection like Regions of Interest. To perform traditional detection methods in grasp detection, they have to implement complex procedures. Nevertheless, our method can simply predict multiple grasp candidates without any other difficult steps as shown in \fig{demo}. Finally, most previous works used low resolution images like 224 $\times$ 224 \cite{Kumra2016Robotic,7139361,zhang2018roi}, our model using 400 $\times$ 400 images to detect robotic grasps. \cite{park2018real}.

Our proposed model first down-samples the image to encode features for detecting robotic grasps and then decodes the features to output pixel-level predictions. We train and evaluation our model on Cornell Grasping Dataset and get high accuracy about 94.42\% which outperforms the network proposed in GG-CNN\cite{Morrison2018Closing}  by  nearly 20\%, and run in our personal computer about 8ms for detecting robotic grasps in one image.

\section{Related Work}
For decades, people have been doing researches on robotic grasps \cite{bicchi2000robotic,shimoga1996robot,sahbani2012overview,bohg2014data,caldera2018review}. Most early works \cite{saxena2007robotic,saxena2008robotic} used human-designed features to represent grasps in images or required the full 3-D model of objects to generate grasps\cite{miller2003automatic,ciocarlie2014towards,herzog2012template}. These methods was popular at that time, but they all faced challenges for non-robust features or lacking of the full 3-D models when used in real world applications. 

Recent years,  learning methods have been proved effective in robotic grasp generation. More and more researchers are trying to use neural network to extract features from images and use these features to detect grasps to be executed by robots.

\subsection{Classification based methods}
Jiang  et al. \cite{5980145} first proposed to use a rectangle representation method to estimate the gripper configuration and the rectangle metric to evaluate a grasp. They carefully designed the grasp features from images and tried to search robotic grasps and then rank them to choose the best one.
Deep learning methods were first applied by Lenz et al.  \cite{Lenz2013Deep} to grasp detection. While the process of generating and selecting robotic grasps is similar to the one used in \cite{5980145}, Lenz et al used sparse auto-encoder to directly extract features from images and achieved the accuracy of 75.6\%. Both methods in  \cite{5980145} and \cite{Lenz2013Deep} are time-consuming because of the exhaustive search in images to generate potential grasps and are unlikely to be used to real-time jobs. Mahler et al. \cite{mahler2017dex} proposed the Dex-Net robotic grasp dataset and used it to train a neural network called GQ-CNN to classify potential grasps using analytic grasp metrics. They sampled antipodal grasps from depth images and utilized deep learning to select the one which is most likely to be successful when picked by a robot. Chun et al. \cite{park2018classification} recently added the spatial transformer network to the pipeline of grasp detection and evaluated on the Cornell Grasp Dataset with the accuracy of 89.60\%. With the spatial transformer network, they were able to provide some partial observation for intermediate grasps. Classification based methods tend to be slow, but the procedure is reasonable.

\subsection{Regression based methods}
Redmon et al. \cite{7139361} first proposed using neural network to directly regress the grasp rectangle parameters from images. By removing the steps of searching potential grasps, the regression method is efficient when compared to the classification method. Their accuracy on the Cornell Grasp Dataset was about 88.0\% with the prediction time of 76ms.
Zhang et al. \cite{zhang2017robust} spent more time on combing rgb features and depth features for accurate grasp detection. They proposed the multi-modal fusion method to regress robotic grasp configurations from RGB-D images and achieved the accuracy of 88.90\% for image-wise split and 88.20\% for object-wise split and the computation time of 117ms. Kumra et al. \cite{Kumra2016Robotic} used ResNet as a feature extractor to detect robotic grasps from RGD images by replacing the blue channel of a image by the depth channel, which increases the accuracy to about 89\%. However, regression based methods tend to output the mean value of ground truth grasps, which may lead to invalid grasps when used.

\subsection{Detection based methods}
Chu et al. \cite{Chu2018Deep,8403246} recently proposed using some key ideas from object detection fields to help generate robotic grasps from images. Their model incorporated a \emph{ grasp region proposal network} to generate candidate regions for later grasp detection and was inspired by Faster R-CNN. With the help of \emph{ grasp region proposal network} and ResNet feature extractor, their network can generate multiple robotic grasps from one image without any other procedure with high accuracy. They evaluated on the Cornell Grasp Dataset and reported the accuracy of 96.0\% for image-wise split and 96.1\% for object-wide split with the prediction time of 120ms. Part et al. \cite{park2018real} proposed use high resolution images with 
360 $\times$ 360 and Multi-Grasp inspired by YOLO to generate robotic grasps. Their method also used ResNet as the feature extractor and combined the idea of anchor boxes, with which they got the accuracy of 96.6\% for image-wise split and 95.4\% for object-wise split with the prediction time of 20ms.

\section{Problem Description}
Given RGB and Depth images of unknown objects, previous works \cite{Kumra2016Robotic,7139361,5980145,park2018classification} all use different ways to generate antipodal robotic grasps from them. The grasp representation they used was proposed by \cite{5980145} and then simplified by \cite{Lenz2013Deep}.

\begin{equation}\label{old} 
g = \left\{ {x,y,\theta,h,w} \right\} 
\end{equation}

This five dimension rectangle representation includes the center of the rectangle $ (x,y) $, the orientation of the rectangle relative to the horizontal axis of the image $ \theta $, the height and width of the rectangle $ (h,w) $.

Recently, Morrison et al.\cite{Morrison2018Closing} proposed the \emph{grasp map} presentation of robotic grasps, which is a fully new idea to deal with the 2-D grasp representation and achieved good results when using neural network to predict robotic grasps with RGB-D images. We follow the representation proposed by them and design a new fully convolution neural network which gives improvement. The grasp representation is:

\begin{equation}\label{new}
g = \left\{ {\textbf{p},\phi,w,q} \right\}
\end{equation}
where $\textbf{p} = (x,y,z)$ is the center position of the gripper, $\phi$ is the rotation angle relative to the horizontal axis of the image plane, $ w $ is the gripper width and $q$ is the grasp quality. The old grasp representation \ref{old} lacks the quality of a grasp, that is we do not know how good a grasp candidate is. We have to do grasp evaluation if there are multiple grasp candidates. However, with the new representation \ref{new}, we can just choose the grasp with highest quality value. 
 
We also assume the 2-D grasp representation can be projected back to 3-D poses executed by robots when we know the camera calibration results.

Robotic grasps can be detected in the depth image $\textbf{I} = \mathbb{R}^{H \times W}$ with height $H$ and width $W$. 
The grasp in image $\textbf{I}$ is represented by

\begin{equation}
 \tilde{g} = \left\{ {\textbf{s},\tilde{\phi},\tilde{w},\tilde{q}} \right\} 
\end{equation}

where $\textbf{s} = (u,v)$ denotes the center point in pixels, $\tilde{\phi}$ denotes the rotation relative to the camera frame, $\tilde{w}$  denotes the gripper width in pixels and $\tilde{q}$ denotes the grasp quality.

The \emph{grasp map} proposed in  \cite{Morrison2018Closing} is
 
\begin{equation}
 \textbf{G} = \left\{ {{\Phi},\textbf{W},\textbf{Q}} \right\}  \in{\mathbb{R}^{3 \times H \times W}} 
\end{equation}

where ${\Phi},\textbf{W},\textbf{Q}$ are each $ \in{\mathbb{R}^{1 \times H \times W}}$ and each pixel contains the $\tilde{\phi},\tilde{w},\tilde{q}$ values respectively.

Like \cite{Morrison2018Closing}, we use neural network to directly generate a grasp $\tilde{\textbf{g}}$ for each pixel in depth image \textbf{I}, which denotes the pixel-wise grasp representation.

\begin{equation}
M(\textbf{I}) = \textbf{G}
\end{equation}

where the map function M can be approximated by deep neural network and then the best grasp can be found by $\tilde{\textbf{g}}^* = \max \limits_{\textbf{Q}} \textbf{G} $.

\section{Approach}
\subsection{Grasp Representation}
In order to compare our model to the GG-CNN, we use all the same grasp representation in Section \uppercase\expandafter{\romannumeral4} of \cite{Morrison2018Closing}.

\begin{figure}[tb] 
\centering 
\subfigure{ \includegraphics[width=8cm]{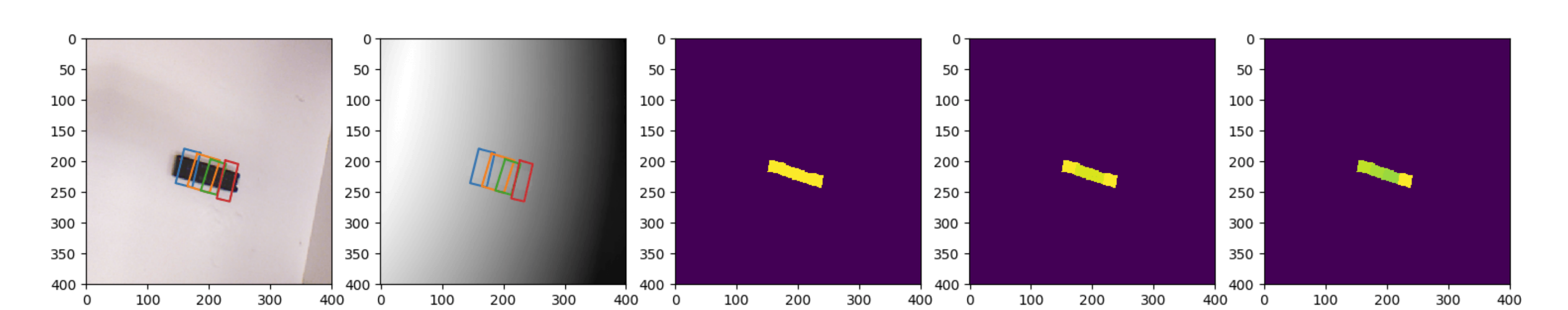} } 
\quad 
\subfigure{ \includegraphics[width=8cm]{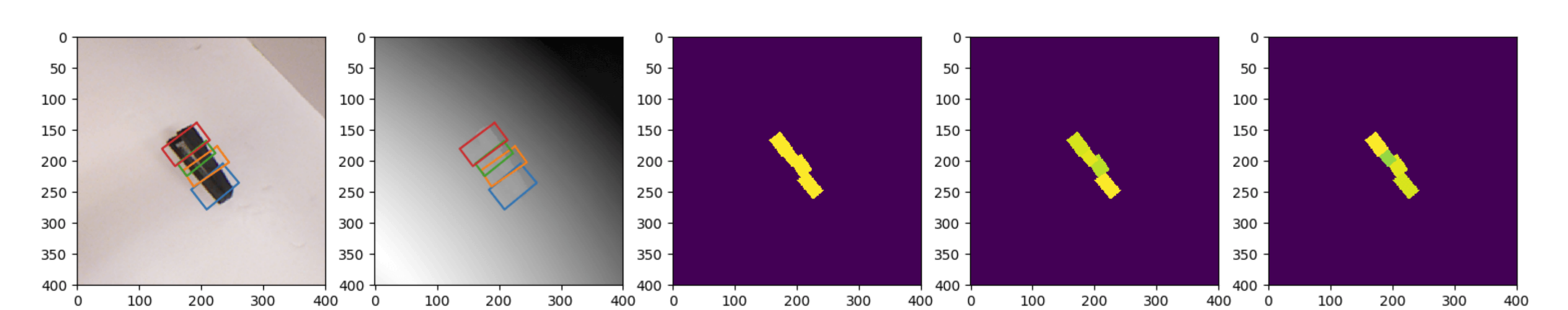} } 
\quad 
\subfigure{ \includegraphics[width=8cm]{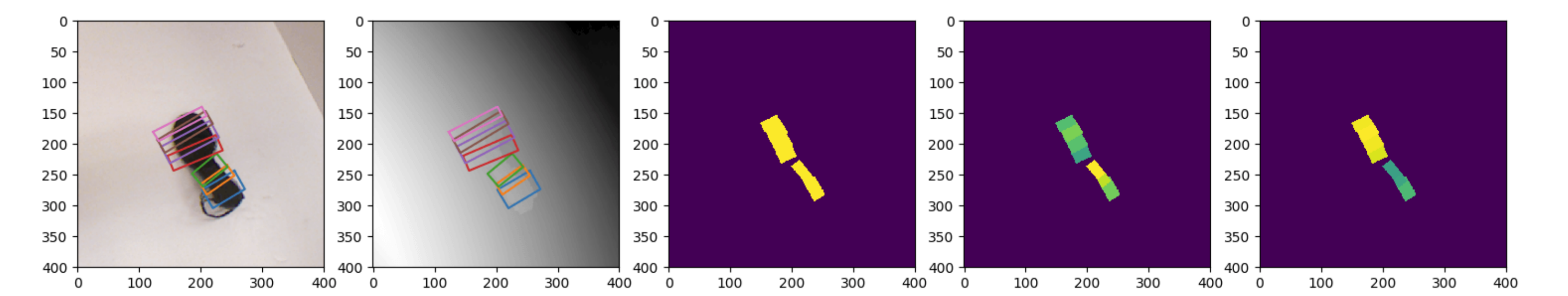} } 
\quad 
\subfigure{ \includegraphics[width=8cm]{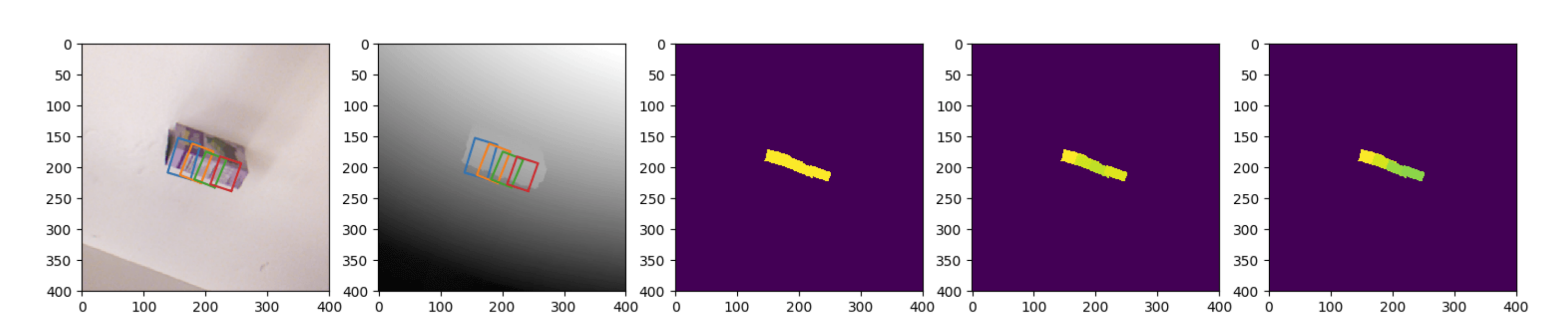} } \caption{Examples from the Cornell Grasp Dataset and their \emph{grasp maps}. From left to right: color image, depth image, grasp position map, grasp angle map, grasp width map.  } 
\label{fig:dataset}
\end{figure}

To build relationships between the rectangle representation of grasps and \emph{grasp maps}, Morrison et al. \cite{Morrison2018Closing} proposed using the center third of the grasp rectangle as an image mask and then using this mask to set corresponding pixel wise properties of robotic grasps. This process is actually doing segmentation to the grasps. For the fact that the position, width and angle values of one robotic grasp is needed, each ground truth positive rectangle will be converted to three small \emph{grasp maps} in pixels. In each \emph{grasp map}, only the region covered by the mask is taken account of, like shown in Fig. \ref{fig:dataset}.

\subsection{Training and Evaluating dataset}
\begin{figure}[tb]
\centerline{\includegraphics[width=8cm]{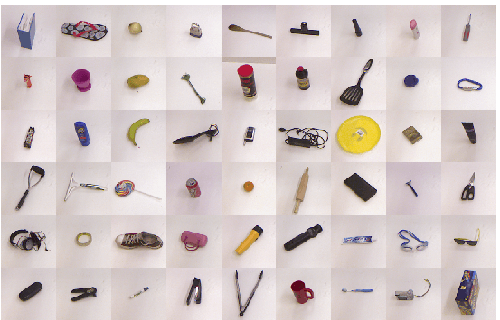}}
\caption{Images from Cornell Grasp Dataset.}
\label{fig:cornell}
\end{figure}

The Cornell Grasp Dataset\cite{Lenz2013Deep}, which shown in \fig{cornell}, contains 885 RGB-D images of real world objects with thousands of positive and negative grasp rectangles.

In order to compare our proposed model with the one in GG-CNN \cite{Morrison2018Closing}, we generate similar dataset by random cropping, zooming and rotating images and corresponding rectangles. However, we find the zooming range$\footnote{https://github.com/dougsm/ggcnn}$ used by Morrison et al. \cite{Morrison2018Closing} may lead to too small objects to be used to evaluate. As a result, we decrease the zooming range to 0.8 and increase the rotation range to 20 degree in order to increase the diversity of grasps. When generating our dataset, we only use the positive grasps and store each robotic grasp representation separately for later usage.

\subsection{Neural Network Architecture}
\begin{figure*} 
\centering 
\subfigure{ \includegraphics[width=18cm,height=10cm]{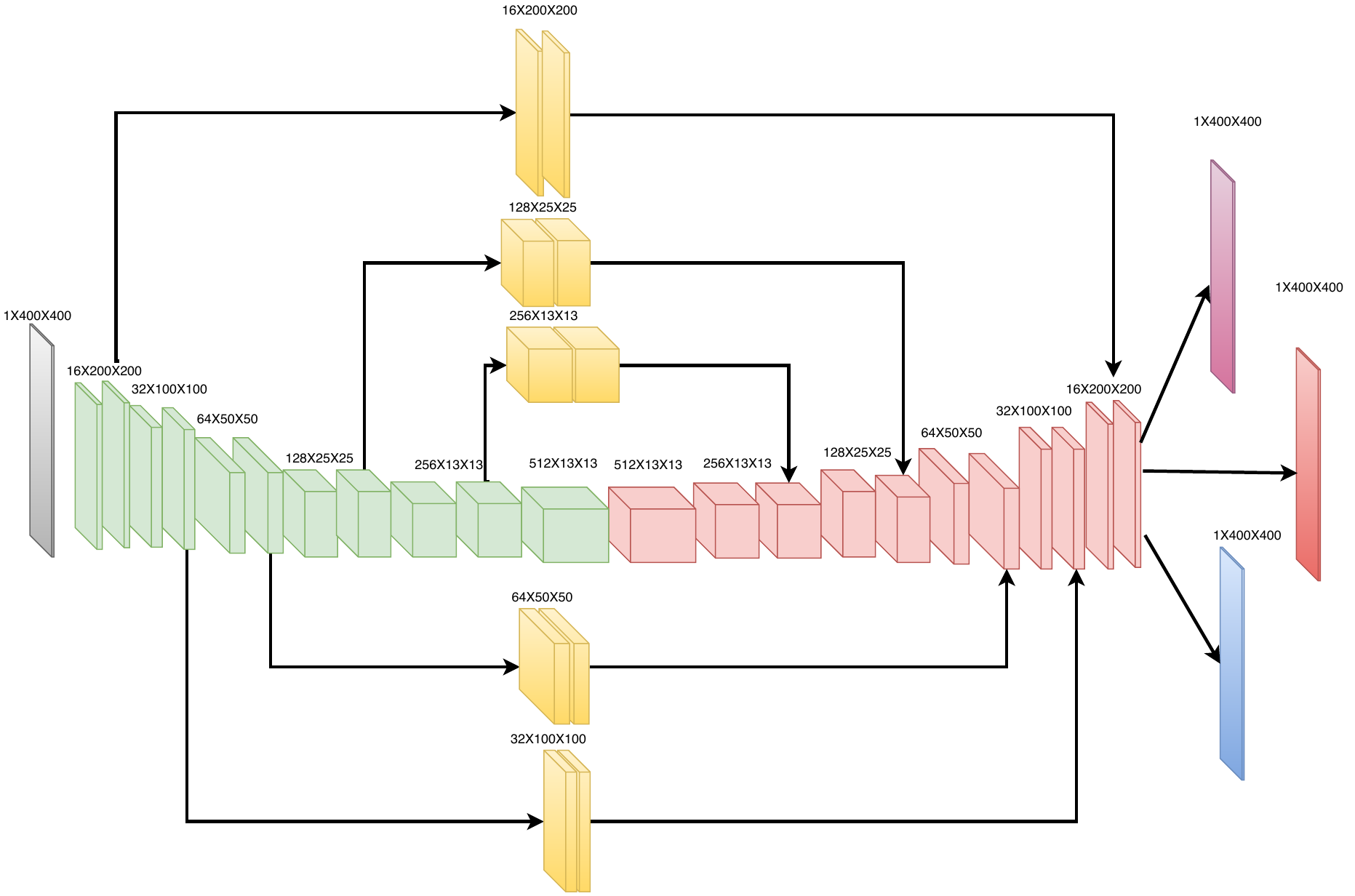} }
\caption{ Proposed fully convolution neural network architecture. }
\label{fig:model}
\end{figure*}

The fully new model we propose to approximate the function M is shown in \fig{model}. 

The pipeline works as one encoder and one decoder. The encoder extracts features for detecting robotic grasps and then the decoder output the pixel-wise grasp parameters. We add Residual Block like connection to the network architecture for the combination of local and global features as well as gradient flows. We use relu as the default activation function to all layers and add batch normalization to each convolution and deconvolution layer. For each pair of convolution or deconvolution layer, the kernel size is the same and we use 9-5-3-3-3-3 for down-sampling part and 3-3-3-5-9-5 for up-sampling part. For all the connections, we use the kernel size of 3 by 3. Our network takes one depth image with size of 400 $\times$ 400 as input and through the fully convolution implementation, three 400 $\times$ 400 \emph{grasp maps} will be predicted to generate robotic grasps. This pixel-wise output then can be used to find the best grasp using $\tilde{\textbf{g}}^* = \max \limits_{\textbf{Q}} \textbf{G} $ and the corresponding angle and width value in the other two \emph{grasp maps}. Moreover, the output can also be used to predict multiple grasp candidates directly because of the pixel-wise implementation. The grasp candidate number actually can be set like those pictures shown in \fig{demo}, and the only change we need to do is to find the expected number of the local maxima in $\textbf{G}$. For $\tilde{g} \in \textbf{G}$, we  can also filter some grasps with low $ \tilde{q}$ value, which might not be so robotic. This method is simple, efficient and powerful when compared to other classification, regression and detection methods.

\section{Experiments}
\subsection{Data Pre-processing}
For the input to our network is the $400 \times 400$ depth image and considering there might be invalid depth values, we do depth in-paint to the image. Before training and evaluating, the grasp width is normalized to be in range [0,1]. All the dimensions are set to the default PyTorch standard\cite{paszke2017automatic}. 

\subsection{Training}
Image-wise split and object-wise split are two default training methods on Cornell Grasp Dataset. For both of the training methods, we do five fold cross validation to better evaluate the performance of our model. For each fold of cross validation, we train the neural network from scratch for 100 epochs and use the batch size of 32. The optimizer we use is the Adam and learning rate is set to 0.001. We also propose to use weighted mean square error as the loss function when training and assign weights according to their importance. Therefore, the total loss can be calculated by

\begin{equation}
\begin{aligned}
  L(\hat{q},\hat{\theta},\hat{w})  =  \frac{1}{2n} [\lambda_q \sum_{u=0}^W \sum_{v=0}^H (\hat{q}_{u,v} - q_{u,v})^2 + \\
 \lambda_\phi \sum_{u=0}^W \sum_{v=0}^H (\hat{\phi}_{u,v} - \phi_{u,v})^2 +  \\
  \lambda_w \sum_{u=0}^W \sum_{v=0}^H (\hat{w}_{u,v} - w_{u,v})^2]   \\ 
\end{aligned}
\end{equation}

where n is the number of training examples; $\hat{q}$,  $\hat{\phi}$,  $\hat{w}$ are the model predictions, $q,\phi,w$ are the ground truth labels; $\lambda_q$, $\lambda_\phi$ and $\lambda_w $ are the weight value for each sub-loss function. When training, we set  $\lambda_q=5$, $\lambda_\phi=3$  and  $\lambda_w=4$ separately.

\subsection{Evaluation}
The standard rectangle metric is used by us to evaluate our model on Cornell Grasp Dataset. A predicted grasp is considered as a valid grasp if it satisfies both of the two conditions:

1) The grasp angle difference between the predicted grasp and ground truth grasp is less than $30^\circ$.

2)The Jaccard index calculated by the predicted grasp and ground truth grasp is greater than 0.25, and the Jaccard index is defined as:

\begin{equation}
\begin{Large}
J(\hat{g},g) = \frac{ \left|\hat{g} \bigcap g \right| }{\left|\hat{g} \bigcup g \right|} 
\end{Large}
\end{equation}

where $\hat{g}$ represents the output prediction of the network and g represents the ground truth label. In fact, the Jaccard index measures how well a prediction matches a ground truth label.

\section{Results}
\begin{figure}[tb] 
\centering 
\subfigure{ \includegraphics[width=3.5cm, height=4cm]{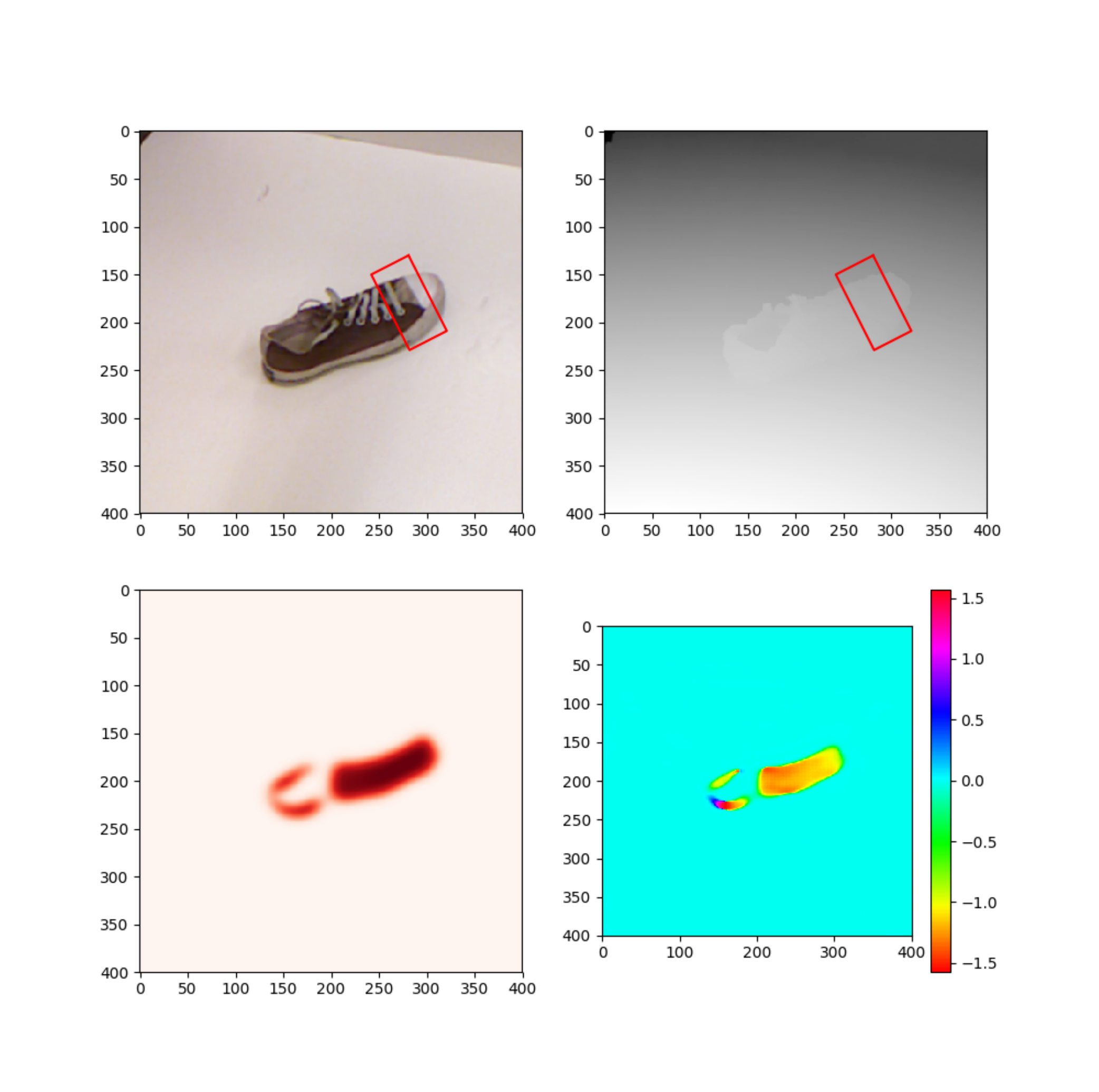} } 
\quad 
\subfigure{ \includegraphics[width=3.5cm, height=4cm]{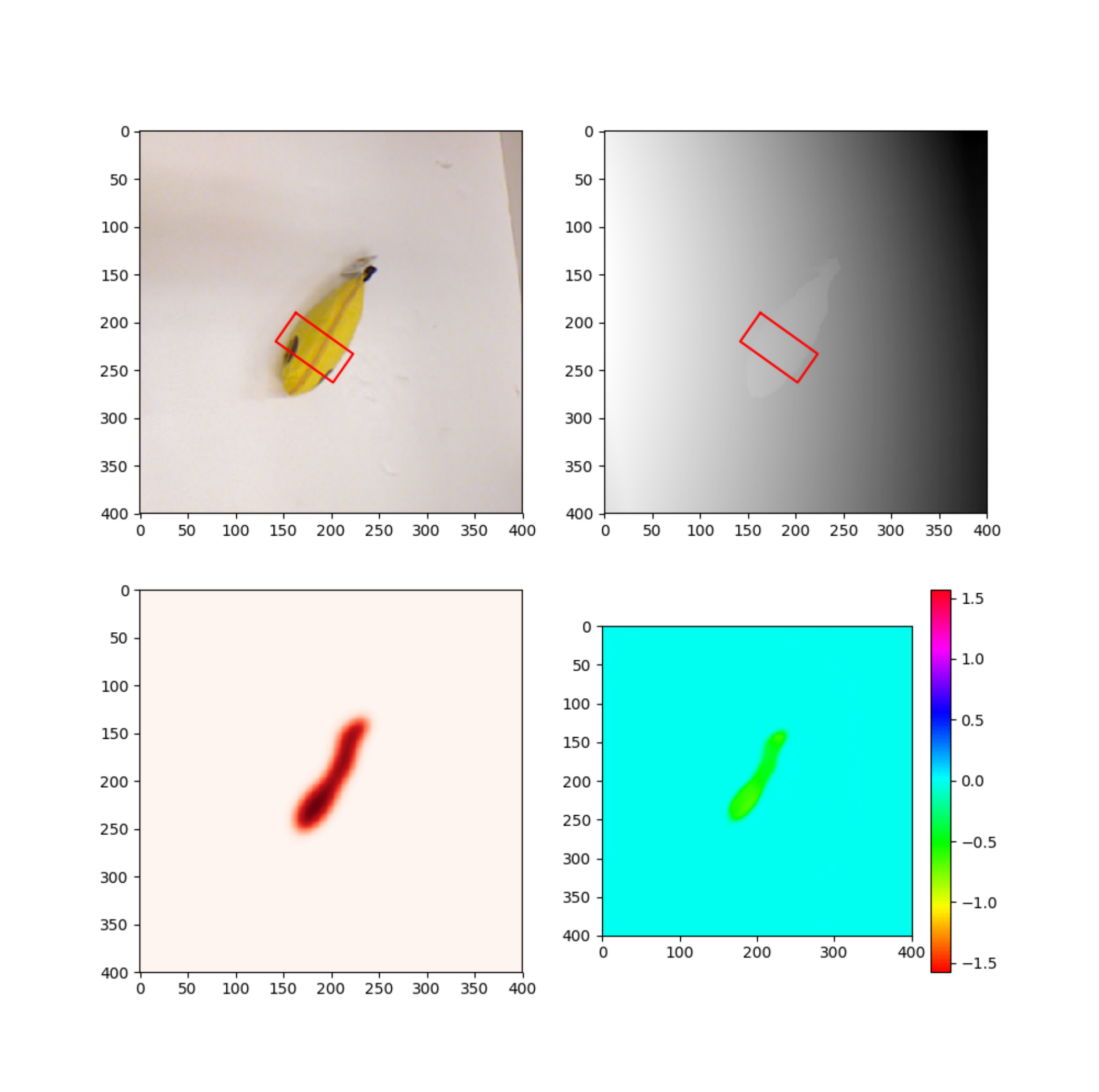} } 
\subfigure{ \includegraphics[width=3.5cm, height=4cm]{fig/Figure_1.pdf} } 
\quad 
\subfigure{ \includegraphics[width=3.5cm, height=4cm]{fig/Figure_2.pdf} } 
\quad 
\caption{ Some correct predictions on Cornell Grasp Dataset. } 
\label{fig:result}
\end{figure}

We implement our model and all related code in PyTorch  and evaluate the performance of our model on the platform with a single GPU (NVIDIA GeForce GTX 1060), a single CPU (Intel i7-8700K 3.7GHz) and 32 GB memory. After doing five fold cross validation, our model finally get the accuracy about 94.42\% for image-wise and 91.02\% for object-wise with the prediction time of only 8ms. \fig{result} shows the image wise training predictions of our model on Cornell Grasp Dataset. For the grasp quality map, the redder the color, the better the quality of a grasp. And for the grasp position map,  the different color stands for different grasp angles.  We compare our results to others in Table \ref{compare}.

\begin{table}[tb] 
\centering 
\caption{ACCURACY ON CORNELL GRASP DATASET}
\begin{tabular}{|c|c|c|c|c|c|c|} 
\hline

\multicolumn{1}{|c}{\multirow{2}*{Model}} & \multicolumn{1}{|c}{\multirow{2}*{Input Size}}
 &  \multicolumn{4}{|c|}{Accuracy} &\multirow{2}*{Time}  \\

\cline{3-6} 
\multicolumn{1}{|c|}{} &\multicolumn{1}{|c|}{}&\multicolumn{2}{c|}{Image} &\multicolumn{2}{|c|}{Object}  &\\ 
\hline

\multicolumn{1}{|c|}{ \multirow{1}*{Redmon et al.\cite{7139361}} }& \multicolumn{1}{|c|}{ \multirow{1}*{$224\times 224$}} &\multicolumn{2}{c|}{88.00\%} &\multicolumn{2}{c|}{87.10\%}&\multirow{1}*{76 ms}\\
\hline 

\multicolumn{1}{|c|}{ \multirow{1}*{Zhang et al.\cite{zhang2017robust}} }& \multicolumn{1}{|c|}{ \multirow{1}*{$224\times 224$}} &\multicolumn{2}{c|}{88.90\%} &\multicolumn{2}{c|}{88.20\%}&\multirow{1}*{117 ms}\\
\hline 

\multicolumn{1}{|c|}{ \multirow{1}*{Kumra et al.\cite{Kumra2016Robotic}} }&\multicolumn{1}{|c|}{ \multirow{1}*{$224\times 224$}} &\multicolumn{2}{c|}{89.21\%} &\multicolumn{2}{c|}{88.96\%}&\multirow{1}*{10 ms}\\
\hline

\multicolumn{1}{|c|}{ \multirow{1}*{Jiang et al.\cite{5980145}} } &\multicolumn{1}{|c|}{ \multirow{1}*{$227\times 227$}} & \multicolumn{2}{|c|}{60.50\%} &\multicolumn{2}{c|}{58.30\%}&\multirow{1}*{-}\\
\hline 

\multicolumn{1}{|c|}{ \multirow{1}*{Lenz et al.\cite{Lenz2013Deep}} }& \multicolumn{1}{|c|}{ \multirow{1}*{$227\times 227$}} &  \multicolumn{2}{c|}{73.90\%} &\multicolumn{2}{c|}{75.60\%}&\multirow{1}*{13.50 sec}\\
\hline

\multicolumn{1}{|c|}{ \multirow{1}*{Chu et al.\cite{8403246}} }&\multicolumn{1}{|c|}{ \multirow{1}*{$227\times 227$}} &\multicolumn{2}{c|}{96.00\%} &\multicolumn{2}{c|}{96.10\%}&\multirow{1}*{120 ms}\\
\hline

\multicolumn{1}{|c|}{ \multirow{1}*{Asif et al.\cite{asif2018graspnet}} }&\multicolumn{1}{|c|}{ \multirow{1}*{$244\times 244$}} &\multicolumn{2}{c|}{90.60\%} &\multicolumn{2}{c|}{90.20\%}&\multirow{1}*{24 ms}\\
\hline 

\multicolumn{1}{|c|}{ \multirow{1}*{Morrison et al.\cite{Morrison2018Closing}} }&\multicolumn{1}{|c|}{ \multirow{1}*{$300\times 300$}} &\multicolumn{2}{c|}{78.56\%} &\multicolumn{2}{c|}{-}&\multirow{1}*{7 ms}\\
\hline

 \multicolumn{1}{|c|}{ \multirow{1}*{Chun et al.\cite{park2018real}} }&\multicolumn{1}{|c|}{ \multirow{1}*{$360\times 360$}} &\multicolumn{2}{c|}{96.60\%} &\multicolumn{2}{c|}{95.40\%}&\multirow{1}*{20 ms}\\
\hline 

\multicolumn{1}{|c|}{ \multirow{1}*{Chun et al.\cite{park2018classification}} }&\multicolumn{1}{|c|}{ \multirow{1}*{$400\times 400$}} &\multicolumn{2}{c|}{89.60\%} &\multicolumn{2}{c|}{-}&\multirow{1}*{23 ms}\\
\hline

\multicolumn{1}{|c|}{ \multirow{1}*{$Ours^*$} }&\multicolumn{1}{|c|}{ \multirow{1}*{$400\times 400$}} &\multicolumn{2}{c|}{94.42\%} &\multicolumn{2}{c|}{91.02\%}&\multirow{1}*{8 ms}\\
\hline

\end{tabular} 

\label{compare}
\end{table}

\begin{figure}[tb] 
\centering 
\subfigure{ \includegraphics[width=3.5cm,height=4cm]{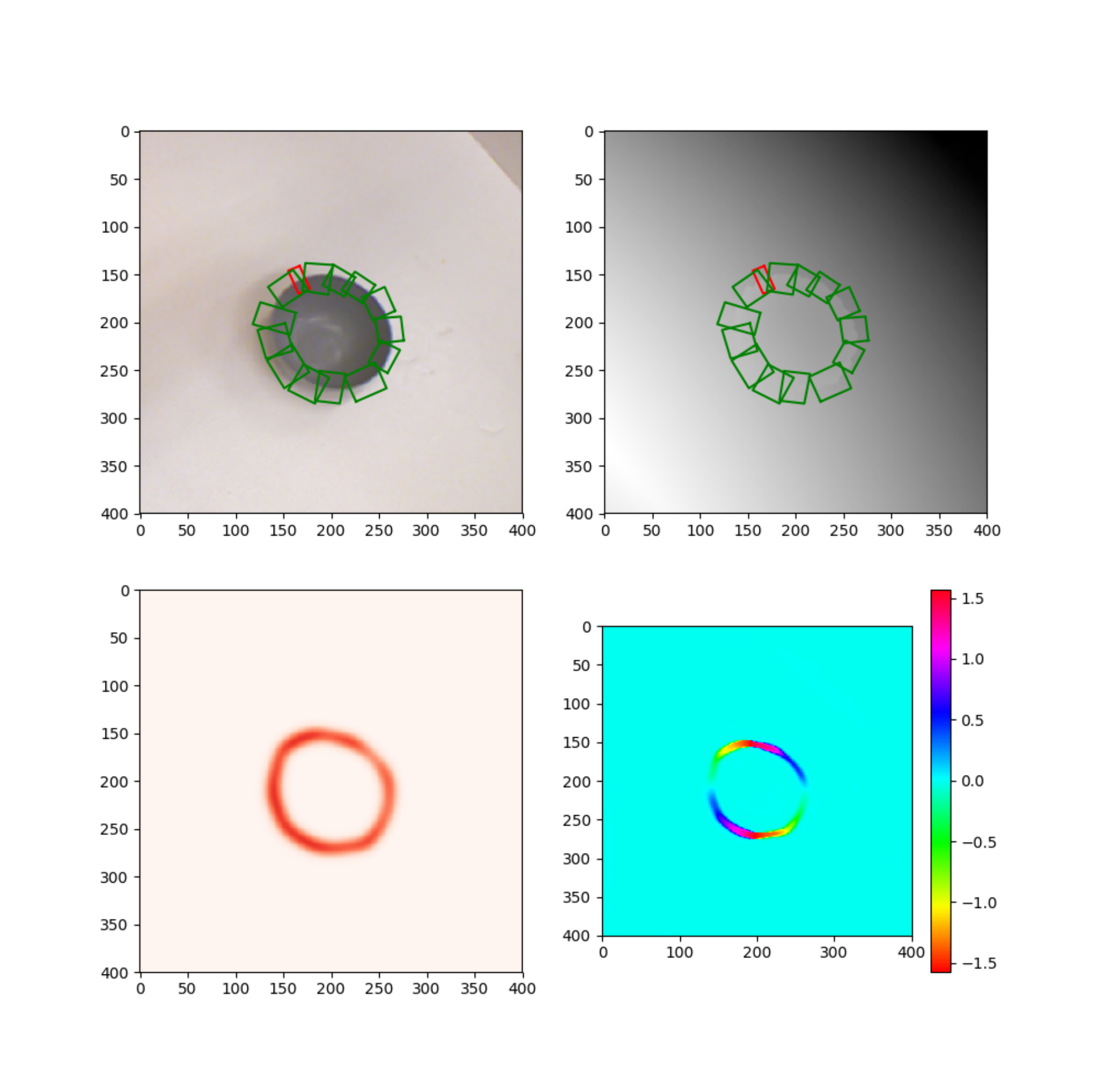} } 
\quad 
\subfigure{ \includegraphics[width=3.5cm,height=4cm]{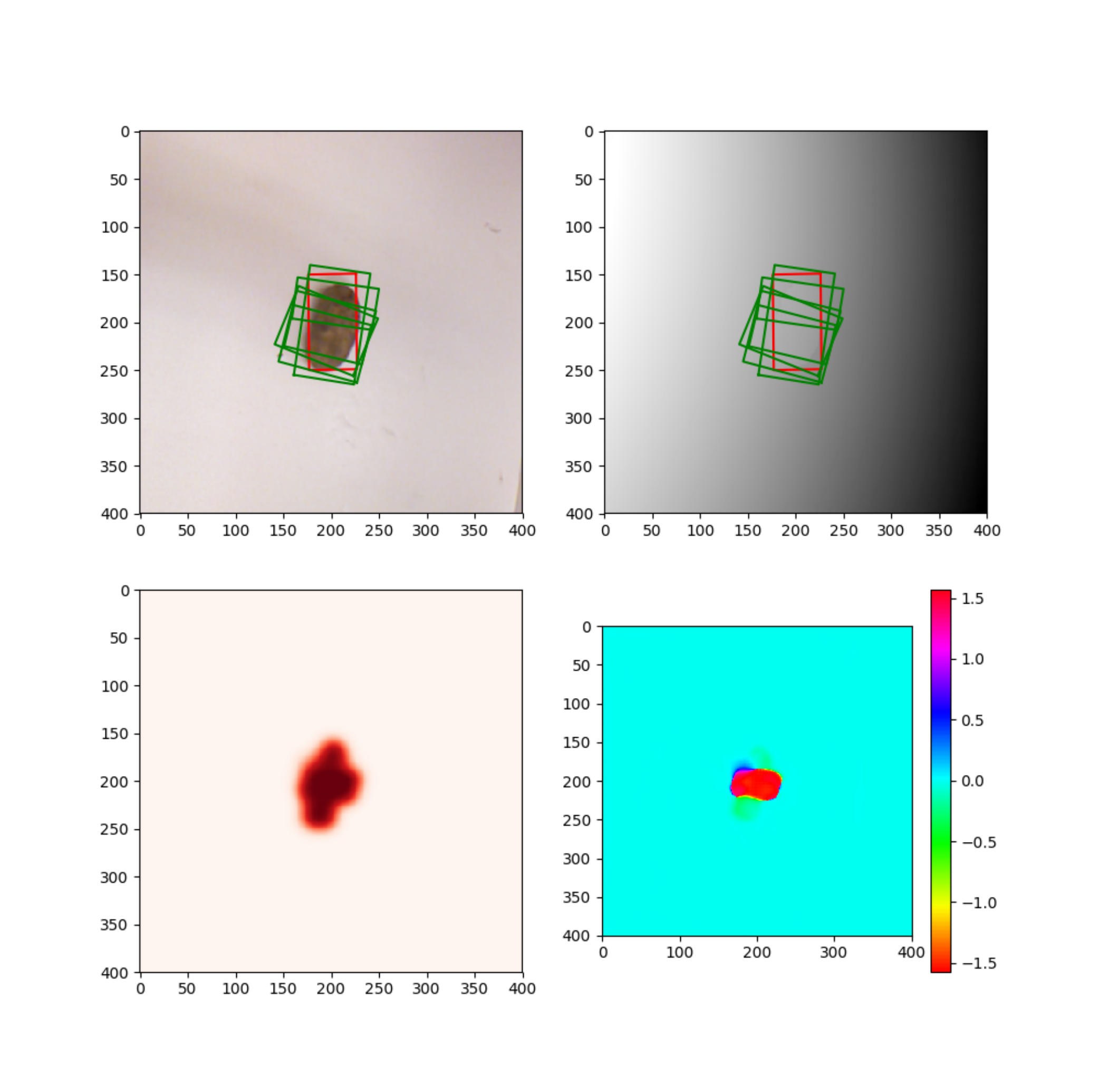} } 
\subfigure{ \includegraphics[width=3.5cm,height=4cm]{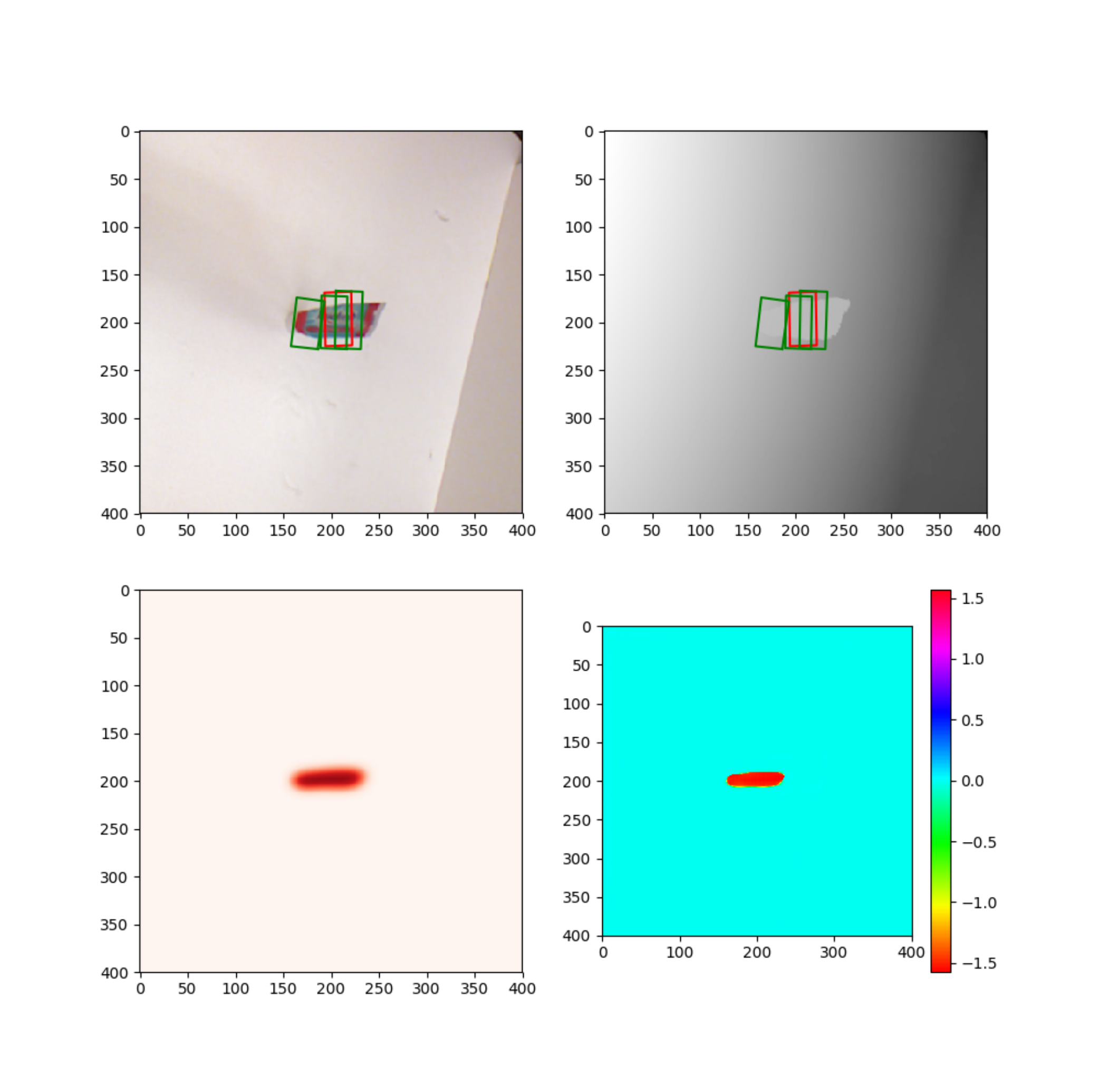} } 
\quad 
\subfigure{ \includegraphics[width=3.5cm,height=4cm]{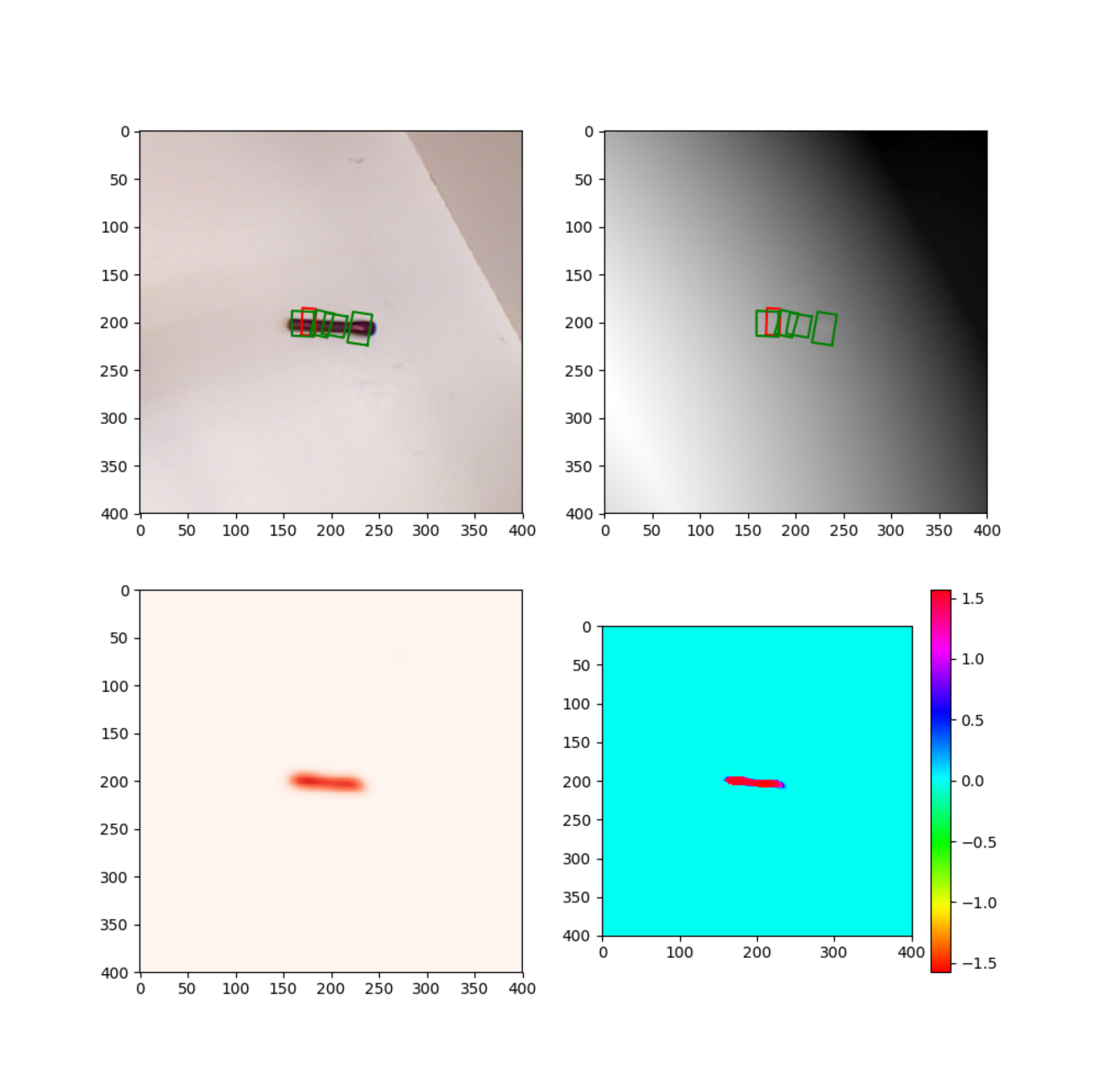} } 
\quad 

\caption{Some false negative predictions. The green ones are the ground truth labels, the red one is the prediction by our network.} 
\label{fig:false negative}
\end{figure}

When evaluating our model, we find our model is able to predict the robust grasps even if it does not see the objects in training set but may be judged incorrect for some new grasps not included in the ground truth labels. These false negative grasps are shown in \fig{false negative}, where the ground truth labels are shown in green and the prediction is shown in red. As a result, the accuracy of our model on Cornell Grasp Dataset might be much higher than the one we report. We also do stricter Jaccard indexes mentioned in \cite{Chu2018Deep} in Table \ref{jaccard}, from which we can see that our model is also able to achieve high accuracy even if the metric is  stricter.

\begin{table}[tb] 
\centering 
\caption{DIFFERENT JACCARD THRESHOLDS RESULT}
\begin{tabular}{|c|c|c|c|c|}
\hline 
Split&0.25&0.30&0.35&0.40\\
\hline  
Image-Wise&94.42\%&92.83\%&90.20\%&85.79\%\\
\hline 
Object-Wise&91.02\%&89.15\%&83.12\%&80.43\%\\
\hline 
\end{tabular}
\label{jaccard}
\end{table}

Also we train and evaluate the GG-CNN \cite{Morrison2018Closing} on the same platform with the Keras implementation and get the accuracy of about 78.56\% with the prediction time 7ms when trained in image-wise split.

\begin{figure}[tb] 
\centering 
\subfigure{ \includegraphics[width=3.5cm,height=4.5cm]{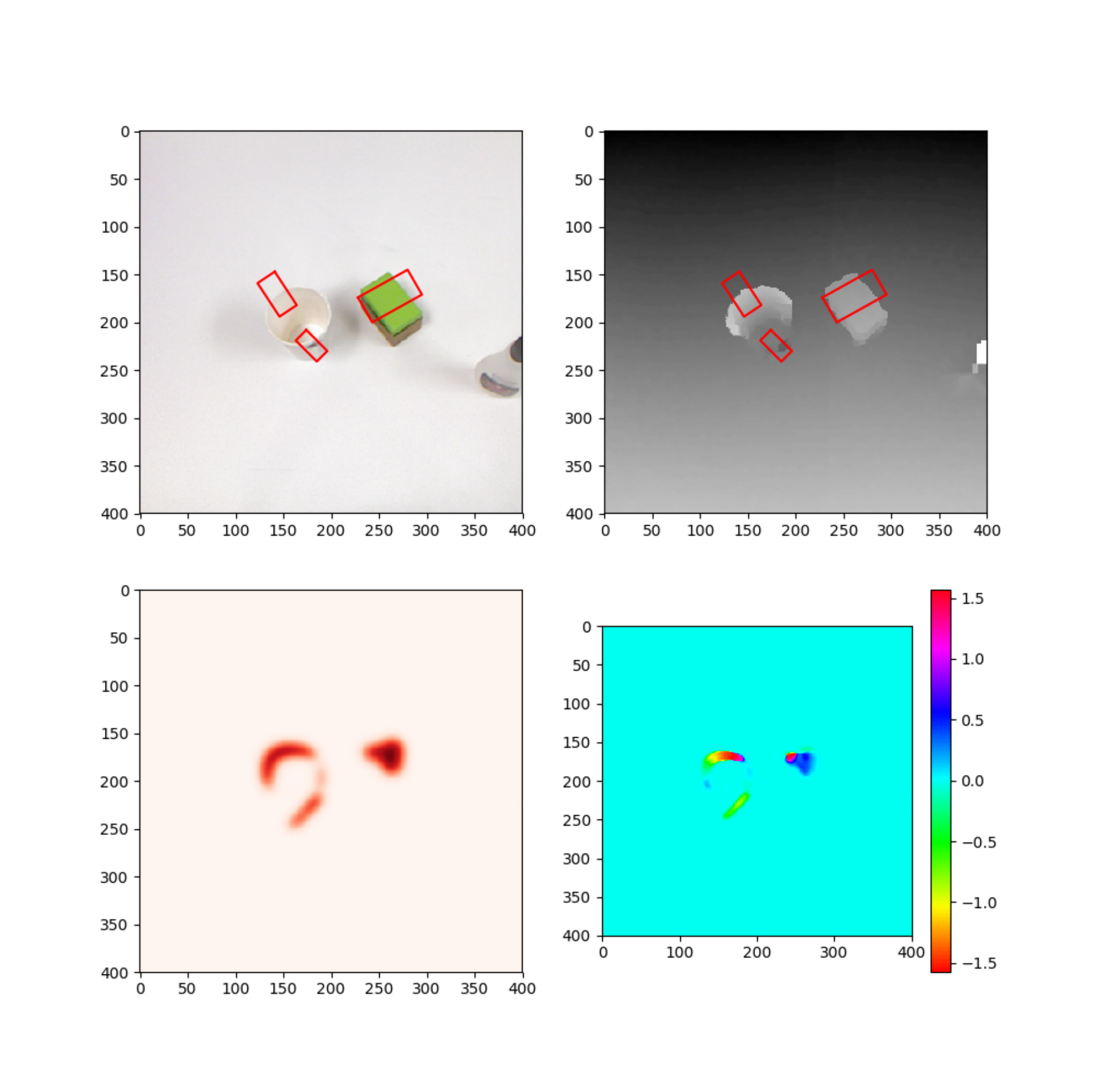} } 
\quad 
\subfigure{ \includegraphics[width=3.5cm,height=4.5cm]{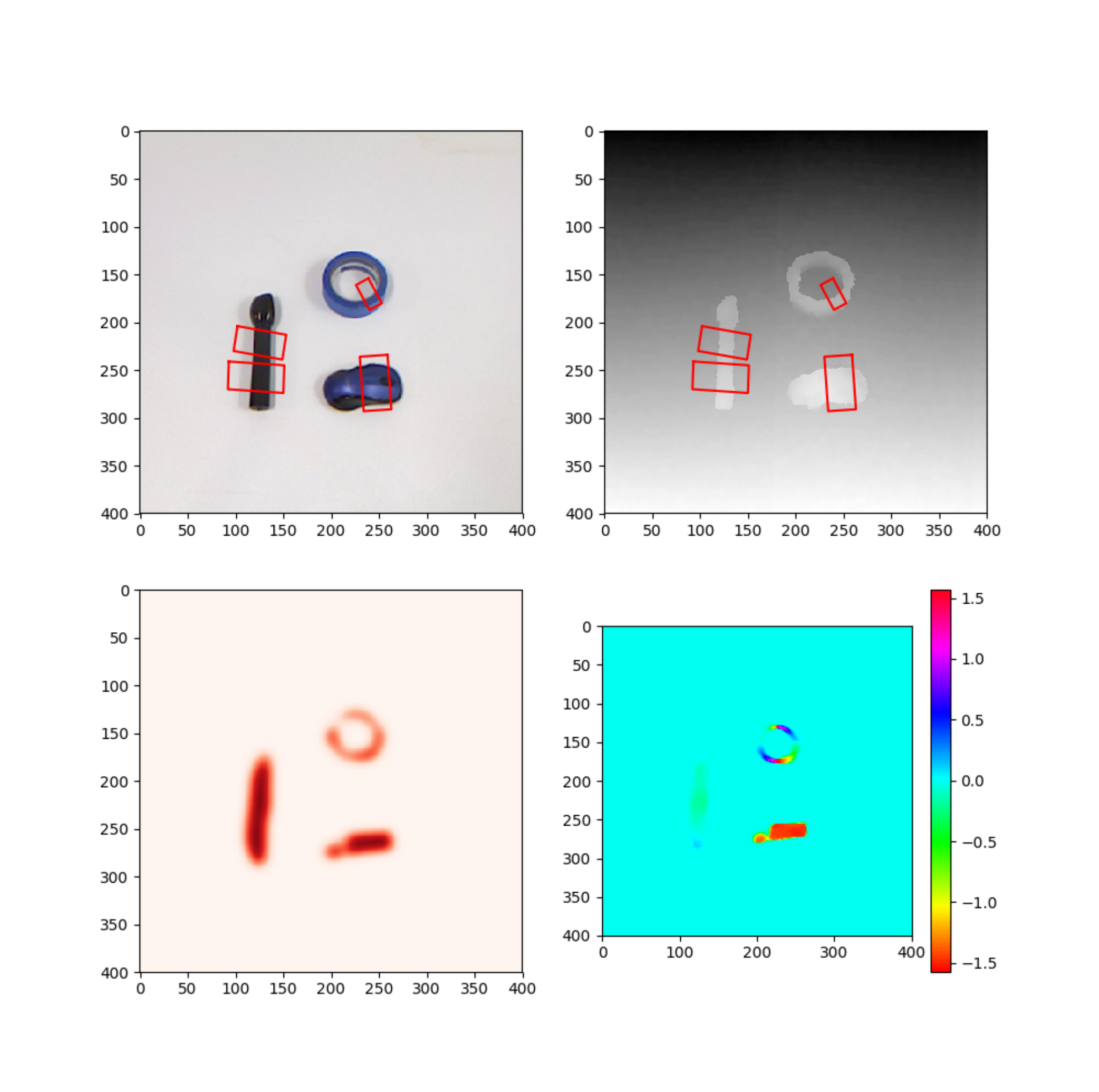} } 
\quad 
\caption{ Predictions on dataset proposed by \cite{Chu2018Deep} with multiple objects } 
\label{fig:georgia}
\end{figure}

For the fact that only one object is contained per image in the standard Cornell Grasp Dataset, Chu at el. \cite{Chu2018Deep} proposed the Georgia Grasp Dataset which contains multiple objects in an image and all the objects are the same as the ones in Cornell Grasp Dataset to evaluate the model performance on multiple objects. We also would like to evaluate the ability of our model to directly generate multiple grasp candidates in images without any other complex procedures used in the detection methods \fig{georgia}. When trying to predict multiple robotic grasps, we set the grasp quality threshold to be 0.5 in order to filter some grasps to have more robotic results. For the fact that we can actually control the number of grasp candidates generated in images, we do not calculate the same false positives per image (FPPI)and Miss Rate mentioned in \cite{Chu2018Deep}. Instead we use the same metric on Cornell Grasp Dataset and do evaluation to each grasp generated in Georgia Grasp Dataset to calculate the whole accuracy when we pick different number of grasps to be generated. When we trying to evaluate the object-wise trained network on the Georgia Grasp Dataset with multiple grasps, we find that our network is not so confident of its predictions and gives low grasp quality $\tilde{q}$ when compared to the image-wise trained network. If we still set the filter threshold to be 0.5, we cannot find so many maxima in the \emph{grasp quality map}. As a result, when we evaluate the object-wise trained model, we set the threshold to be 0.2 in order to increase the number of predictions. However, there might be more grasps which are not so robotic or even not on the object if we do so. There is a trade-off between the filter threshold and number of predictions to be generated. Our final results is shown in Table \ref{different number}.

\begin{table}[tb] 
\centering 
\caption{DIFFERENT NUMBER OF PREDICTIONS ACCURACY}
\begin{tabular}{|c|c|c|c|c|c|}
\hline 
\# of predictions&1&2&3&4&5\\
\hline  
Image-Wise&89.52\%&87.16\%&85.36\%&84.79\%&83.84\%\\
\hline  
Object-Wise&90.36\%&89.64\%&87.60\%&83.93\%&80.06\%\\

\hline

\end{tabular}
\label{different number}
\end{table}

\section{Conclusion}
In this research, a new fully convolution neural network is presented and it generates robotic grasps by using high resolution depth images. Our propose model encodes the origin input images to features and then decode these features to generate robotic grasp properties for each pixel. It demonstrates well performance on the Cornell Grasp Dataset. Unlike other methods for generating multiple grasp candidates through neural network, the pixel-wise implementation can directly predict multiple grasp candidates through one forward propagation and we can use these pixel-wise results to filter some grasps and even control the number of grasps to be generated. The trained model size is only about 35 MB and the computation time 8ms is fast enough to perform real-time robotic applications with high accuracy.

\bibliography{ref}
\bibliographystyle{IEEEtran}

\end{document}